%% file: neurips_2025.tex
\documentclass{article}

 \usepackage[preprint]{neurips_2025}

\usepackage[utf8]{inputenc} 
\usepackage[T1]{fontenc}    
\usepackage{hyperref}       
\usepackage{url}            
\usepackage{booktabs}       
\usepackage{amsfonts}       
\usepackage{nicefrac}       
\usepackage{microtype}      
\usepackage{xcolor}         
\usepackage{amsmath}
\usepackage{enumitem}
\usepackage{graphicx}
\usepackage{amssymb}
\usepackage{xspace}
\usepackage{multirow}
\usepackage{wrapfig}
\usepackage{subcaption}
\usepackage[ruled,vlined]{algorithm2e}
\usepackage{titletoc}
\usepackage{wrapfig}

\title{From Image Generation to Infrastructure Design: a Multi-agent Pipeline for Street Design Generation}

\author{
 \textbf{Chenguang Wang}\textsuperscript{1,2},
 \textbf{Xiang Yan}\textsuperscript{3},
 \textbf{Yilong Dai}\textsuperscript{3},
 \textbf{Ziyi Wang}\textsuperscript{4},
 \textbf{Susu Xu}\textsuperscript{1}\\
 \textsuperscript{\rm 1}Johns Hopkins University~~~~
 \textsuperscript{2}Stony Brook University\\
 \textsuperscript{3}University of Florida~~~~
 \textsuperscript{4}University of Maryland 
 \\
 \texttt{chenguang.wang@stonybrook.edu}~~~~\texttt{susuxu@jhu.edu}
}

\begin{document}

\maketitle

\begin{abstract}
\vspace{-1em}
Realistic visual renderings of street-design scenarios are essential for public engagement in active transportation planning. Traditional approaches are labor-intensive, hindering collective deliberation and collaborative decision-making. While AI-assisted generative design shows transformative potential by enabling rapid creation of design scenarios, existing generative approaches typically require large amounts of domain-specific training data and struggle to enable precise spatial variations of design/configuration in complex street-view scenes. We introduce a multi-agent system that edits and redesigns bicycle facilities directly on real-world street-view imagery. The framework integrates lane localization, prompt optimization, design generation, and automated evaluation to synthesize realistic, contextually appropriate designs. Experiments across diverse urban scenarios demonstrate that the system can adapt to varying road geometries and environmental conditions, consistently yielding visually coherent and instruction-compliant results. This work establishes a foundation for applying multi-agent pipelines to transportation infrastructure planning and facility design.
\end{abstract}

\input{introduction}

\input{method}

\input{result}

\input{conclusion}

\bibliography{custom}
\bibliographystyle{plain}

\appendix

\clearpage
\input{related_work}

\clearpage
\input{supplement}

\clearpage
\input{prompt}

\end{document}

%% file: introduction.tex
\vspace{-1.5em}
\section{Introduction}
\vspace{-1em}

\textit{Cycling is an environmentally friendly mode of transportation that also offers co-benefits such as promoting personal health and reducing traffic congestion ~\citep{fishman2015dutch, fishman2016cycling, brand2021climate, liu2024planning}. However, bicycle infrastructure development often requires extensive stakeholder consultations, during which road users (e.g., cyclists, drivers, and pedestrians) articulate their needs and concerns.} To facilitate these deliberations, visual renderings of proposed street design scenarios are widely used in practice as tools for collective reflection and collaborative decision-making. Traditionally, these visuals are created with graphic design software (e.g., Adobe Photoshop and SketchUp)~\citep{al1999using, bosselmann1999livable, sheppard2001guidance}.  While effective, these tools are time-consuming and demand specialized expertise, making it difficult to customize and dynamically adjust street design images in response to various user feedback, thereby hindering agile scenario iteration and limiting their utility in dynamic public engagement contexts that involve complex trade-offs in allocating road space~\citep{bickerstaff2002transport, lofgren2020designing, hancock2024challenges, lawlor2023stakeholders}.

Recent advances in Generative AI (GenAI), particularly image-generation models, demonstrate significant potential to support scenario ideation and facilitate collaborative decision-making across domains such as industrial design, architecture, and site planning.~\citep{onatayo2024generative, fang2023role, wang2025generative, fu2025towards, hancock2024challenges, lawlor2023stakeholders, kapsalis2024urbangenai}. Existing GenAI-based scenario design has leveraged post-training methods on domain-specific imagery~\citep{regmi2019cross, lu2020geometry, deng2024streetscapes, gu2025text2street}, which necessitates large, curated datasets and significant computational resources. More recently, the state-of-the-art models such as GPT-image-1~\citep{openai_4o_image_generation, qian2025gie} have made it feasible to apply off-the-shelf systems directly to design scenarios without task-specific retraining. These models provide strong text-to-image and image-to-image capabilities, enabling the rapid creation of immersive, design-oriented bicycle-infrastructure visualizations from urban imagery, particularly street-view imagery. However, research on customizing these models for street infrastructure design, particularly at the site level, remains largely underexplored. 
Key limitations include: (i) inadequate reasoning about spatial and relational structure within visual inputs; (ii) semantic misinterpretation of user instructions; (iii) weak adherence to complex instructions that with multiple constraints specified; and (iv) inconsistent outputs and occasional hallucinations ~\citep{qian2025gie, vemishetty2025towards, shirakawa2024noisecollage}. These limitations underscore that stand-alone image generation is insufficient for bicycle-infrastructure scenario design, pointing to the need for a more structured framework that situates state-of-the-art models within a more comprehensive workflow.

To leverage cutting-edge GenAI models for bicycle infrastructure scenario design while addressing the existing limitations, we propose a multi-agent system built on a state-of-the-art image generation backbone, GPT-image-1~\citep{openai_4o_image_generation}. Given a user-defined prompt and street-view imagery, the generative pipeline produces realistic bicycle facility design scenarios via four specialized agents that can tackle each of the four limitations discussed above: (1) a \textit{Locator Agent} that generates contextually accurate descriptions of bike-lane positions using Multimodal Large Language Models (MLLMs), helping image generation model capture spatial relations; (2) a \textit{Prompt Optimization Agent} that refines user prompts by integrating illustrative references along with the Locator’s contextual descriptions, thereby reducing semantic misinterpretation. (3) a \textit{Design Generation Agent} that decouples geometric and design-pattern constraints via a cascading generation, yielding multiple candidate scenario designs. (4) an \textit{Evaluation Agent} that reranks candidate designs via CLIP similarity to a reference layout and conducts a binary compliance check with reasoning MLLMs, surfacing the most instruction-aligned outputs. Experimental results on street-view imagery collected from diverse road contexts show that our pipeline consistently generates realistic, instruction-aligned, and spatially coherent designs. This enables rapid creation of street-design scenarios and supports collective reflection and collaborative decision-making in bicycle infrastructure planning and design.

The contributions of this work are threefold:
\vspace{-0.5em}
\begin{itemize}
\item We extend the applicability of generative AI in urban planning by integrating state-of-the-art image-generation models for bicycle infrastructure design.
\vspace{-0.4em}
\item We develop a multi-agent system that generates street infrastructure configurations with high spatial accuracy and contextual relevance, while ensuring compliance with planning guidelines.
\vspace{-0.4em}
\item We design a pipeline that streamlines the design workflow, reducing complexity, expertise requirements, and time cost in scenario generation.
\end{itemize}

%% file: method.tex
\section{Methodology}

\begin{figure}
    \centering
    \includegraphics[width=0.95\linewidth]{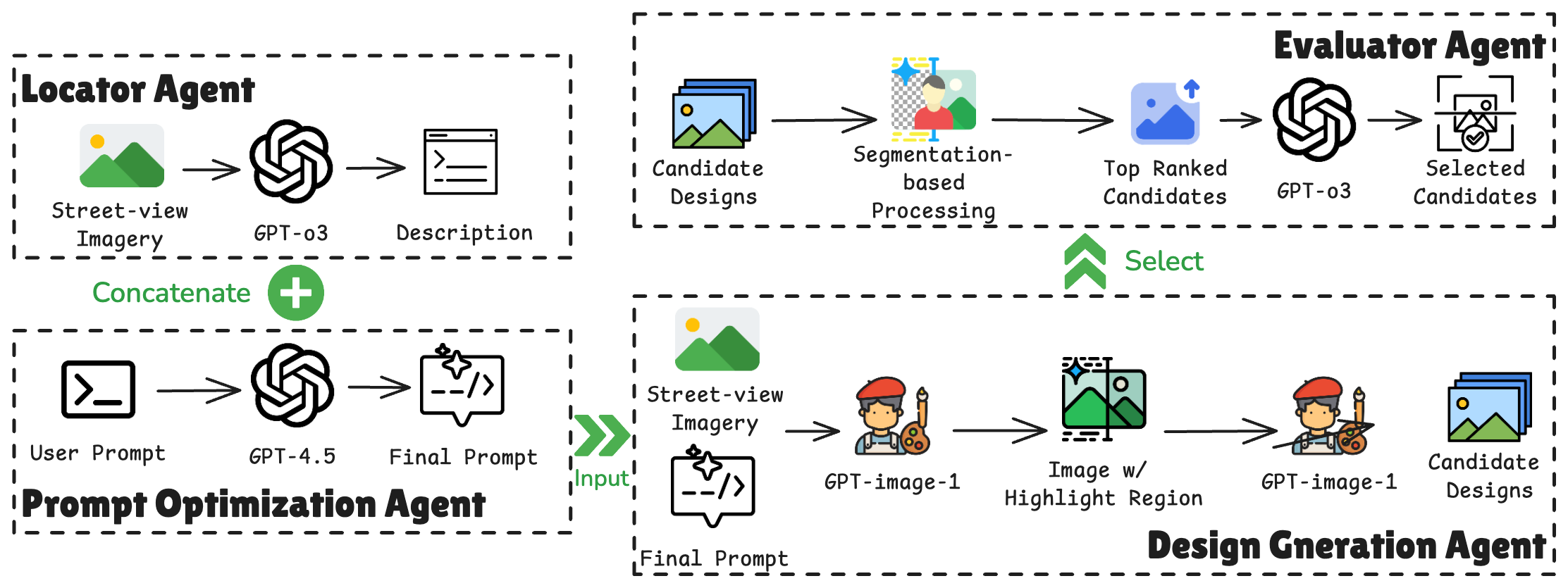}
    \caption{\textbf{Overview of our multi-agent system for bicycle-infrastructure design.}
    The system comprises a Locator, Prompt Optimization, Design Generation, and Evaluation agent that processes street-view imagery to generate bike-lane designs. Green arrows and plus signs denote intermediate operations on agents' output within the workflow.}
    \label{fig:main_pipeline}
\end{figure}

This section presents our multi-agent system for bicycle infrastructure design. As we discuss in detail in Section 3.1, the pipeline takes a street-view image and a user-specified design request as inputs, generates multiple bicycle-lane design candidates by editing the input image, and selects a final design that best satisfies the user-specified design requirements (Fig.~\ref{fig:main_pipeline}). To demonstrate scalability and broad applicability, we sampled Google Street View imagery from Washington DC across diverse road environments, obtained via the Google Street View API (see Section 3.2).

\subsection{Method: A multi-agent system for bicycle infrastructure design}

According to the previous discussion, design-grade bicycle-infrastructure visualization from a single street-view image poses a constraint-rich editing problem that is poorly served by single-pass prompting. Small geometric or contextual errors, e.g., drifting lane boundaries, inconsistent curbside offsets, misplaced separators, or unintended changes to adjacent street elements, can fundamentally alter the meaning of a proposed intervention and undermine deliberation-grade use. Moreover, bicycle-lane treatments typically require a \emph{sequence} of dependent, localized edits (identify the curbside region, remove or repurpose parking, introduce buffers and separators, repaint markings, and harmonize lighting/materials) while preserving a fixed camera viewpoint; mistakes introduced early can propagate and compound in later steps. These characteristics motivate an editing-centric, verifiable workflow that makes intermediate states explicit and attaches checking/repair to each stage. Following agentic visual systems that decompose long-horizon visual tasks into modular roles ~\citep{gupta2023visual, hang2025cca, sun2025marmot, li2025mccd}, we implement a role-specialized multi-agent pipeline: the \textit{Locator Agent} grounds the scene by extracting curbside geometry and proposing an intervention region; the \textit{Prompt Optimization Agent} translates user intent into a constraint-emphasizing editing specification; the \textit{Design Generation Agent} executes the intervention by producing a diverse candidate set via a two-step cascading strategy (region isolation followed by lane-treatment rendering); and the \textit{Evaluator Agent} performs selection and verification by re-ranking candidates with lane-focused representations and checking compliance against the optimized specification.

Our pipeline operationalizes these requirements through four design choices that map directly to the above agents and target the dominant failure modes in street-scene editing: (i) \textbf{explicit spatial grounding} (Locator) to anchor lane boundaries and insertion regions to stable physical references and to filter scenes that do not support unambiguous curbside edits; (ii) \textbf{constraint clarity and prompt robustness} (Prompt Optimization) via a structured template and in-context exemplars, reducing prompt sensitivity and limiting unintended changes to non-target elements; (iii) \textbf{execution fidelity for compositional edits} (Design Generation) via two-step cascading generation, which improves locality and reduces geometric drift compared with direct single-pass compositional prompting while providing diverse candidates for downstream selection; and (iv) \textbf{verifiable selection} (Evaluator) using lane-focused masking for similarity re-ranking together with an LLM-based compliance check, prioritizing candidates that satisfy explicit requirements rather than merely appearing plausible. Together, these components enable rapid, viewpoint-consistent bicycle-lane scenario iteration without costly task-specific retraining of the image-generation backbone, while making failure detection and repair an explicit part of the workflow.

Compared with prompt-only or single-pass editing workflows, our pipeline explicitly decomposes the task into spatial grounding, instruction formalization, staged execution, and post-hoc verification. Each module targets a distinct failure mode in street-scene editing: the Locator Agent reduces ambiguity in lane placement, the Prompt Optimization Agent reduces under-specified or internally inconsistent instructions, the Design Generation Agent mitigates compositional drift through highlight-guided execution, and the Evaluator Agent filters stochastic but noncompliant generations. This decomposition is inspired by recent agentic visual systems, but is adapted here to the geometry-sensitive and deliberation-oriented setting of bicycle-facility design, where even small errors in boundary placement or background preservation can substantially change the meaning of a proposed intervention.

\subsubsection{Locator Agent.}
Preliminary experiments and prior studies suggest that current image generation models often struggle to accurately interpret and preserve the spatial configuration of street-scene elements in street-view imagery, which can lead to unreliable placement and geometry of bicycle infrastructure when prompted directly~\citep{qian2025gie}. To address this limitation, we designed a \textit{Locator Agent}. Given a street-view image, the agent produces a structured, spatially grounded description of bicycle-infrastructure-relevant geometry. When a bike lane is present, the agent describes its key attributes, including markings/paint patterns, approximate width, and relative position to stable reference elements such as the curb, sidewalk edge, parking lane, or travel lanes, with explicit cues for left/right boundaries. Formally, let $x\in\mathbb{R}^{H\times W\times 3}$ be an RGB street-view image. The locator agent is a multimodal mapping
\begin{equation}
(\hat{s},\hat{\mathcal{R}},\hat{\mathcal{B}},\hat{\mathbf{a}})=L(x),
\end{equation}
where $\hat{s}\in\{0,1\}$ indicates suitability, $\hat{\mathcal{R}}=\{\hat{r}_j\}_{j=1}^J$ is a set of grounded reference polylines (e.g., curb/sidewalk edge, lane lines), and $\hat{\mathcal{B}}=(\hat{b}_L,\hat{b}_R)$ are the left/right bike-lane boundary polylines in image coordinates (either observed when a lane exists, or proposed as an insertion region otherwise); $\hat{\mathbf{a}}$ contains attributes such as marking/paint type and relative placement w.r.t.\ $\hat{\mathcal{R}}$. A simple width descriptor reported by the locator agent is the mean boundary separation:
\begin{equation}
\hat{w}=\mathbb{E}_{t\sim\mathcal{U}[0,1]}\left[\left\lVert \hat{b}_L(t)-\hat{b}_R(t)\right\rVert_2\right].
\end{equation}
 When no bike lane is present, the image is \emph{not} discarded; instead, the agent identifies a plausible insertion region by referencing the same stable geometric cues (e.g., curbside alignment and lane-line/edge references) to support downstream addition of a new bike lane with minimal ambiguity. Images are excluded only when the scene is not suitable for bicycle-lane retrofit or addition (e.g., high-speed/limited-access roadway contexts or views lacking clear curbside/lane reference lines), because such cases do not support unambiguous and realistic design synthesis. In our experiments, our locator agent is powered by GPT-o3, a state-of-the-art reasoning multimodal model~\citep{OpenAI_o3_o4mini_system_card}.

\subsubsection{Prompt Optimization Agent.}
Prior work has shown that synthesized image quality depends strongly on prompt formulation, and that in-context prompt generation can substantially improve results~\citep{hao2023optimizing, wang2023context}. Motivated by these findings, we designed a \textit{Prompt Optimization Agent} to translate user requirements and the Locator output into a robust image-editing prompt. We began by manually drafting candidate prompts and comparing different description formats to identify a prompt structure that the image generation backbone can follow reliably. Our exploratory experiments indicate that a structured template yields the most consistent outputs; this template includes an overall lane design description, detailed boundary specifications grounded in stable scene references (left/right boundaries), and explicit constraints that prevent unintended changes to non-target elements. Based on this template, we prepared several high-quality exemplars for in-context learning, integrated user-specific instructions, and prompted a GPT-4.5 model to automatically generate the final image-generation prompt (see Appendix~\ref{sec:prompt}). The generated prompt is then combined with the structured, scene-grounded descriptions produced by the Locator Agent (and any additional constraints), ensuring that downstream generation receives clear, precise, and contextually anchored instructions.

\subsubsection{Design Generation Agent.}
Building on evidence that image generation models may fail to faithfully render all specified elements in complex, compositional prompts~\citep{ghosh2023geneval, huang2023t2i, huang2025t2i}, we designed the \textit{Design Generation Agent} to improve robustness via a two-step cascading strategy. 
In Stage 1, the model generates an intermediate highlighted image rather than the final bicycle-lane design. Specifically, it renders a colored block in the designated intervention area (green in our implementation), which serves as a coarse visual anchor for the subsequent edit. This intermediate output is not intended to represent semantically meaningful lane content. Rather, it marks the spatial region to be modified. The highlighted block is embedded directly in the image and is passed together with the image to Stage 2, allowing the pipeline to separate the spatial grounding of the edit region from the final rendering of the lane treatment.

In Stage 2, the model takes the highlighted intermediate image together with the optimized prompt and transforms the marked region into the final scenario-specific bicycle-lane intervention, such as a standard marked lane, a buffered lane, or a colored-surface lane. The prompt explicitly specifies that the highlighted green region denotes the target area for editing and that all surrounding scene elements should be preserved. This includes roadway texture, lighting, shadows, curbs, sidewalks, parked vehicles, and lane markings outside the intervention area. All generations were produced using the GPT-image-1 model through the Responses API, with output size fixed at 1024 $\times$ 1024, quality is correspondingly set to medium automatically, and PNG is used as the output format. Other low-level controls commonly available in some image-generation systems, such as temperature, editing strength, or random seed, are not exposed as user-configurable parameters in this API and therefore cannot be explicitly controlled. For each scenario, we generated 10 candidates using the same optimized prompt and the same highlighted intermediate image. The resulting variation across candidates arose from the model’s inherent stochasticity and was typically limited to local rendering details, such as lane width, separator spacing, and marking style, rather than the global placement of the intervention.

To ensure that the Stage-1 output provides a valid basis for downstream editing, we perform a lightweight verification step before proceeding to Stage 2. Specifically, the intermediate image is checked for the presence of the designated green region and for approximate alignment with the intended intervention area. This verification combines a simple color-based check with brief human inspection. If the highlighted region is missing or clearly misplaced, Stage 1 is rerun. In cases where a Stage-2 candidate fails to follow the highlighted region and produces substantial spatial deviation or off-target edits, the result is subsequently flagged as noncompliant by the Evaluator Agent and excluded from final selection.

\subsubsection{Evaluator Agent}

Recent work reports that redundant environmental content (e.g., vehicles, pedestrians, and buildings) can undermine embedding-based similarity assessment by introducing noise into representations such as CLIP embeddings~\citep{radford2021learning, li2024context}. To mitigate this problem, we introduce a segmentation-based preprocessing stage that isolates bicycle-infrastructure regions before similarity computation. We manually annotated 1,060 images drawn from both real street-view scenes and model-generated outputs. The generated samples were collected before downstream screening so that the training set would include both satisfactory and unsatisfactory cases. Using these annotations, we fine-tuned a YOLO-v11 instance segmentation model~\citep{yolo11_ultralytics}. The dataset was divided into training, validation, and test sets in a 7:1:2 ratio. On the validation set, the fine-tuned model achieved an mAP@50 of 91.9\%, with precision and recall of 99.5\% and 90.9\%, respectively. These results indicate that the predicted masks are sufficiently accurate for downstream lane-focused preprocessing and candidate re-ranking. The resulting model produces binary masks that isolate bicycle-infrastructure regions, while non-relevant areas are replaced with a uniform color. This preprocessing suppresses extraneous visual content while preserving lane geometry and appearance.

After segmentation, we compute CLIP ViT-B/32 embeddings on the isolated bicycle-infrastructure regions and use cosine similarity to re-rank generated candidates against a designated reference design. Concretely, the predicted bicycle-infrastructure mask is applied to each candidate image and to the corresponding reference image, while all non-relevant background regions are replaced with black in a preprocessing step. The processed images retain the original 1024 $\times$ 1024 resolution so that lane geometry and relative spatial layout are preserved even as irrelevant environmental content is suppressed. We then compute cosine similarity between the processed candidate image and the processed reference design and use this score as a coarse ranking signal. In this stage, higher CLIP similarity generally indicates a closer match to the target lane configuration after background interference has been reduced. Only the top three candidates under this ranking are advanced to the next verification stage.

The final decision is made by an MLLM-based compliance check rather than by CLIP similarity alone. Each of the top three CLIP-ranked candidates is provided to GPT-o3~\citep{OpenAI_o3_o4mini_system_card}, together with its corresponding optimized prompt and, when applicable, the designated reference image, for binary compliance evaluation. The model is instructed to act as a strict evaluator and determine whether the candidate image depicts a bicycle lane on the right side of the roadway that satisfies all required design features specified in the prompt. Minor visual deviations, such as small differences in perspective, lighting, or width, are treated as acceptable only when all required design elements remain clearly present and recognizable. The model outputs a binary suitability judgment indicating whether the candidate should be accepted or rejected. Detailed evaluation prompts are provided in Appendix~\ref{sec:prompt}. More specifically, the compliance check evaluates whether the generated image satisfies the scenario-specific structural and design constraints encoded in the optimized prompt. These include whether the bicycle lane is located on the right side of the roadway, whether the left and right boundaries match the prescribed scenario-specific configuration, whether required protection elements or separators are present and correctly positioned when specified, whether lane surface color, symbols, and buffer or hatching treatments are consistent with the intended design, and whether non-target scene content remains materially unchanged. We treat the design-specific requirements as hard constraints because they determine whether the generated image still represents the intended bicycle-lane scenario. Preservation of non-target scene content is treated as a realism constraint that guards against off-target modification. This second-stage verification ensures that the final selected design is not only visually similar to the reference pattern but also compliant with the explicit structural and semantic requirements of the target scenario.

By default, only the top three CLIP-ranked candidates are sent to the compliance checker. If none of them passes the compliance evaluation, we discard the current candidate pool and regenerate a fresh set of 10 candidates in Stage 2 using the same highlighted intermediate image. The regenerated pool is then re-ranked by CLIP similarity and evaluated again. This fallback strategy avoids selecting a visually similar but noncompliant result and instead relies on renewed stochastic generation when the initial candidate set fails to produce an acceptable design. In terms of computational cost, the segmentation and CLIP-based re-ranking stages are lightweight and run on a single NVIDIA RTX A6000 GPU for segmentation inference and embedding computation. These stages are substantially less expensive than the downstream LLM-based verification stage. In our implementation, the main latency arises from GPT-based compliance checking in the Evaluator Agent, which typically requires about 60-90 seconds per scenario, depending on the number of candidate images reviewed. Consequently, the end-to-end runtime is dominated by the second-stage verification step rather than by segmentation or similarity computation.

\subsection{Street View Image Collection}
\label{sec:sup}

To ensure that the proposed approach is applicable across diverse roadway environments, we sampled 150 road segments and intersections in Washington, D.C., considering factors such as the presence and type of existing bicycle facilities, speed limit, and surrounding neighborhood context. The sampled locations included both sites where bicycle infrastructure was already present and sites without bicycle facilities but with roadway layouts that could plausibly accommodate new bike-lane interventions. This sampling strategy was intended to cover a broad range of roadway conditions rather than a narrow subset of favorable examples.

For each sampled coordinate, we queried the Google Street View API from nearby viewpoints and retrieved static street-view images from multiple horizontal viewing angles while keeping pitch and field of view fixed to maintain a consistent eye-level perspective. This procedure yielded 600 raw candidate images in total. All images were retrieved at a resolution of 1024 $\times$ 1024 to match the downstream image-generation backbone. We then conducted a dedicated quality-control stage in which human reviewers examined the retrieved views for each location and selected the single most suitable image for downstream use. Specifically, the reviewers prioritized images that (1) clearly showed the roadway and curbside region, (2) preserved usable geometric references such as curbs, sidewalks, or lane boundaries, (3) revealed existing bicycle-lane markings when present, (4) provided a forward-looking viewpoint suitable for subsequent bicycle-lane editing, and (5) avoided severe occlusion from vehicles, vegetation, signage, or shadows. The final curated image set, therefore, reflects both broad roadway diversity and explicit human screening effort, providing consistent and contextually appropriate inputs for all downstream stages of the pipeline.

%% file: result.tex
\section{Experiments}
\label{sec:experiments}

\subsection{Design scenarios}
\label{sec:design_scenarios}
Based on common bikeway configurations in street design practice, we define eight representative bicycle-lane design scenarios (DS1--DS8). These scenarios cover standard marked lanes, buffered lanes that increase separation from motor traffic, and treatments with enhanced visibility (e.g., colored surfacing). Table~\ref{tab:boundaries_by_pattern} summarizes the boundary specifications used in our experiments, and Table~\ref{tab:boundaries_by_pattern} provides detailed definitions with visual examples.

\begin{table*}[h]
\centering
\scalebox{0.85}{
\begin{tabular}{c|l|l}
\toprule
\textbf{Design Scenario} & \textbf{Left boundary} & \textbf{Right boundary} \\
\midrule
1 & No buffer; direct adjacency to moving lane & No buffer; direct adjacency to parked cars \\
2 & No buffer; direct adjacency to moving lane & 3 ft white-painted buffer \\
3 & No buffer; direct adjacency to moving lane & 1.5 ft buffer with bollards \\
4 & No buffer; direct adjacency to moving lane & 1.5 ft buffer with armadillo lane dividers \\
5 & No buffer; direct adjacency to moving lane & No buffer; direct edge (no separator) \\
6 & 3 ft white-painted buffer & No buffer; direct edge (no separator) \\
7 & 1.5 ft buffer with bollards & No buffer; direct edge (no separator) \\
8 & 1.5 ft buffer with armadillo lane dividers & No buffer; direct edge (no separator) \\
\bottomrule
\end{tabular}
}
\caption{\textbf{Design scenario specifications used in our experiments.}}
\label{tab:boundaries_by_pattern}
\end{table*}

\subsection{Evaluation protocol}
\label{sec:eval_protocol}
MLLM-based evaluation of generative outputs is widely adopted in recent benchmarks and studies~\citep{gu2024survey}. In this work, we evaluate our multi-agent bicycle-infrastructure pipeline with a stepwise, human-in-the-loop protocol coupled with a final design-quality assessment framework. This protocol is inspired by prior work advocating staged generation with expert oversight and end-outcome metrics such as realism and instruction adherence~\citep{he2025generative}, and is adapted to the specific requirements of bicycle-lane design and strict background preservation.

\paragraph{Human-in-the-loop review at critical stages.}
We incorporate expert oversight at four decision points to reduce error accumulation across agents. During location-description generation, experts verify that the produced description correctly captures bicycle-infrastructure-relevant geometry while respecting the existing road layout (e.g., parking zone boundaries and adjacent travel lanes), and they minimally edit the description when necessary. During prompt optimization, experts check that design objectives and constraints are translated into verifiable clauses (e.g., explicit background-preservation constraints) and iterate until the prompt is both effective for generation and acceptable under design guidelines. During the highlight-region step, experts verify spatial precision and request regeneration when the highlighted region leaks into the base scene. Finally, during candidate selection, the Evaluator Agent nominates the best design from the candidate pool, while experts independently select their preferred output; disagreements trigger targeted upstream revisions (location description, prompt, and/or highlight region) followed by re-generation and re-selection. This staged review-and-iterate protocol keeps expert control at points where errors most strongly affect downstream design realism and compliance.

\paragraph{Final design-quality assessment.}
For the final output (i.e., the agent-selected and expert-approved design), we conduct human evaluation along two axes that directly target failure modes common in street-scene editing: \textit{visual fidelity} and \textit{instruction compliance}. These criteria explicitly capture hallucinations and subtle lane--background inconsistencies that generic image similarity metrics may miss.

\textit{Visual fidelity} measures whether the edited bicycle infrastructure appears realistic and is integrated into the original scene without unintended modifications outside the designated region. Raters assign 1--5 Likert scores on three criteria: (1) lane plausibility (appropriate width, smooth curvature, continuity, and connectivity), (2) scene integration (consistent perspective, shading/shadows, reflections, and material/texture blending), and (3) background preservation (no edits beyond the intended lane region and no semantic drift in roads, sidewalks, or street furniture). We report criterion-wise scores and a composite visual-fidelity score computed from these criteria. The rubric targets recurrent generative failure modes, including geometric artifacts (jagged edges, broken continuity), photometric/texture inconsistencies (incorrect shadows, tiling, color cast), perspective mismatch, improper occlusions (markings painted over vehicles/pedestrians), and background drift.

\textit{Instruction compliance} evaluates whether the output satisfies the optimized prompt. We translate each optimized prompt into a checklist of hard constraints (e.g., lane width, boundary type, marking style, and required protection elements) and soft constraints (e.g., ``do not alter non-target background''). For each checklist item, raters mark satisfied/unsatisfied, yielding a binary compliance vector and an overall compliance rate; we additionally collect a global 1--5 Likert judgment of prompt adherence. Hallucinated or out-of-spec elements (e.g., spurious barriers/crosswalks, invented curb geometry, contradictory markings) are counted as violations, and missing required features are treated as non-compliance.

To align the evaluation with expert decision-making, we further summarize the two-axis rubric into an acceptance-based accuracy metric. A case is counted as correct if it (i) meets a pre-specified visual-fidelity acceptance threshold (including a strict background-preservation condition) and (ii) satisfies all hard instruction constraints, while meeting a minimum level of soft-constraint compliance defined by the checklist. We report accuracy as the proportion of test cases that satisfy these acceptance criteria.

\section{Results}
\label{sec:results}

\subsection{Main results}
\label{sec:main_results}

\begin{figure}
\centering
\includegraphics[width=1\textwidth]{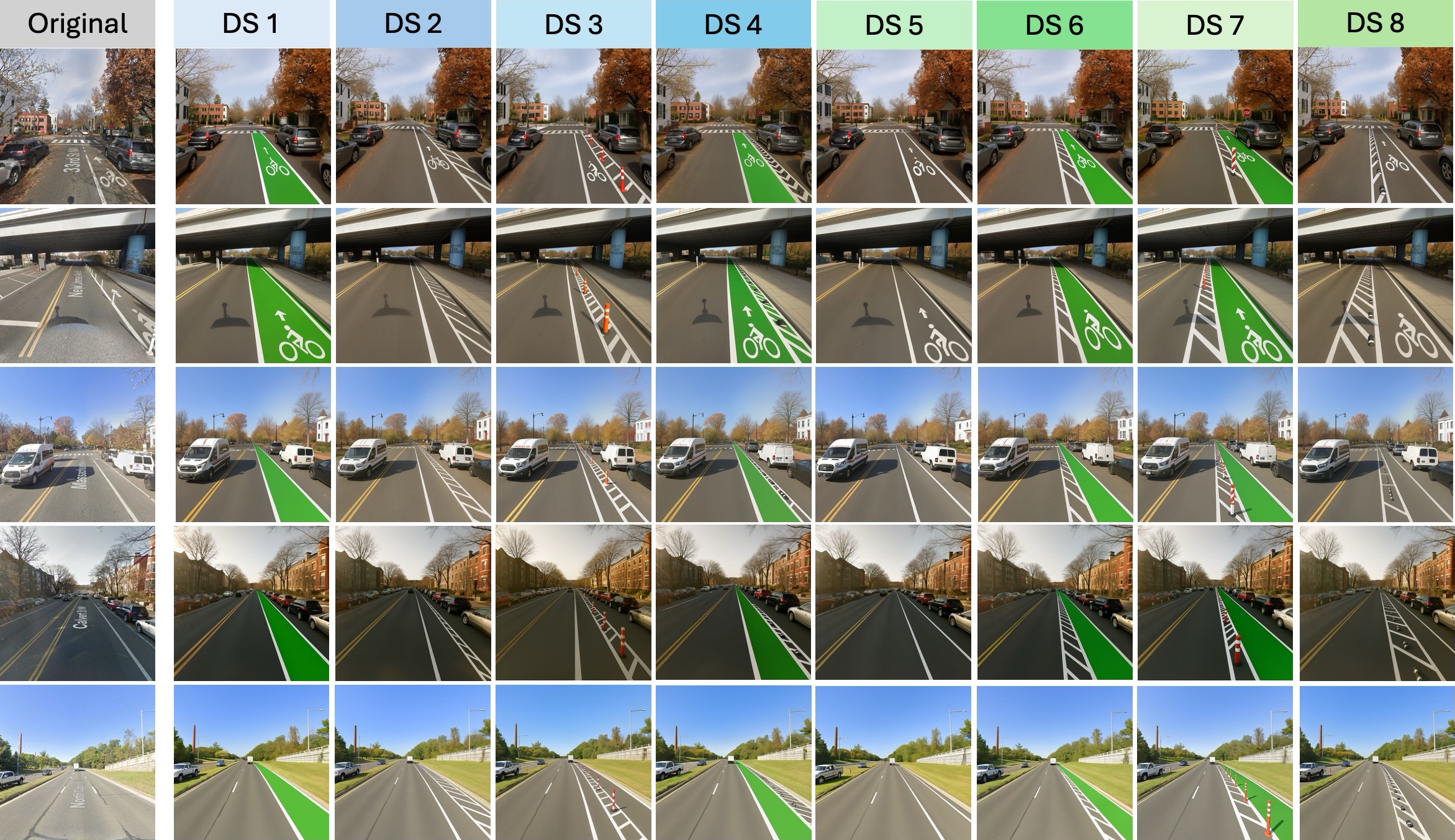}
\caption{\textbf{Generated bicycle-lane designs across diverse urban contexts.} Each row shows the original street-view scene (left) and eight variations generated by our multi-agent pipeline, one per predefined design scenario (DS1--DS8).}
\label{fig:main_results}
\end{figure}

Figure~\ref{fig:main_results} presents representative outputs produced by our multi-agent pipeline across diverse street-view contexts. Each row shows the original scene and eight generated variants corresponding to DS1--DS8. The results indicate that the pipeline can embed different lane patterns into heterogeneous environments (e.g., dense urban streets, suburban corridors, and complex intersections) while maintaining plausible alignment with roadway geometry and preserving the overall scene. Even under visually challenging conditions such as cluttered backgrounds and partial occlusions, the synthesized lanes remain visually distinct and contextually integrated.

In addition to typical urban contexts, we include a small set of deliberately challenging roadway scenes (e.g., highway-like or otherwise unfavorable contexts) as \textit{counterfactual stress tests}. These cases are not intended to imply that bicycle facilities are universally appropriate in such settings; rather, they are used to probe whether the pipeline can maintain localization, background preservation, and prompt compliance under harder geometric and semantic conditions.

\begin{table}[h]
\centering
\scalebox{0.9}{
\begin{tabular}{c|cccccccc}
\toprule
\textbf{Design Scenario} & 1 & 2 & 3 & 4 & 5 & 6 & 7 & 8 \\
\midrule
\textbf{Eval Acc. (\%)} & 95.5 & 96.5 & 97.0 & 95.5 & 96.0 & 95.5 & 97.0 & 96.5 \\
\bottomrule
\end{tabular}
}
\caption{\textbf{Accuracy of the Evaluator Agent in selecting the most suitable candidate.}}
\label{tab:discriminator_accuracy}
\end{table}

\subsection{Evaluator Agent performance}
\label{sec:evaluator_results}
To assess whether the Evaluator Agent can reliably select a design that adheres to the final optimized prompt, we conduct a human evaluation on a held-out street-view test set that was not used during training. Table~\ref{tab:discriminator_accuracy} reports scenario-wise selection accuracy. Across all eight scenarios, the Evaluator Agent achieves accuracy above 95\%, indicating robust selection under stochastic candidate variation and supporting its role as a stabilizing component for final design choice.

\subsection{Ablation studies}
\label{sec:ablation}
We further assess the contribution of each agent by removing individual components and comparing outputs to the full pipeline under otherwise matched inputs. 

\begin{figure}
    \centering
    \includegraphics[width=0.75\linewidth]{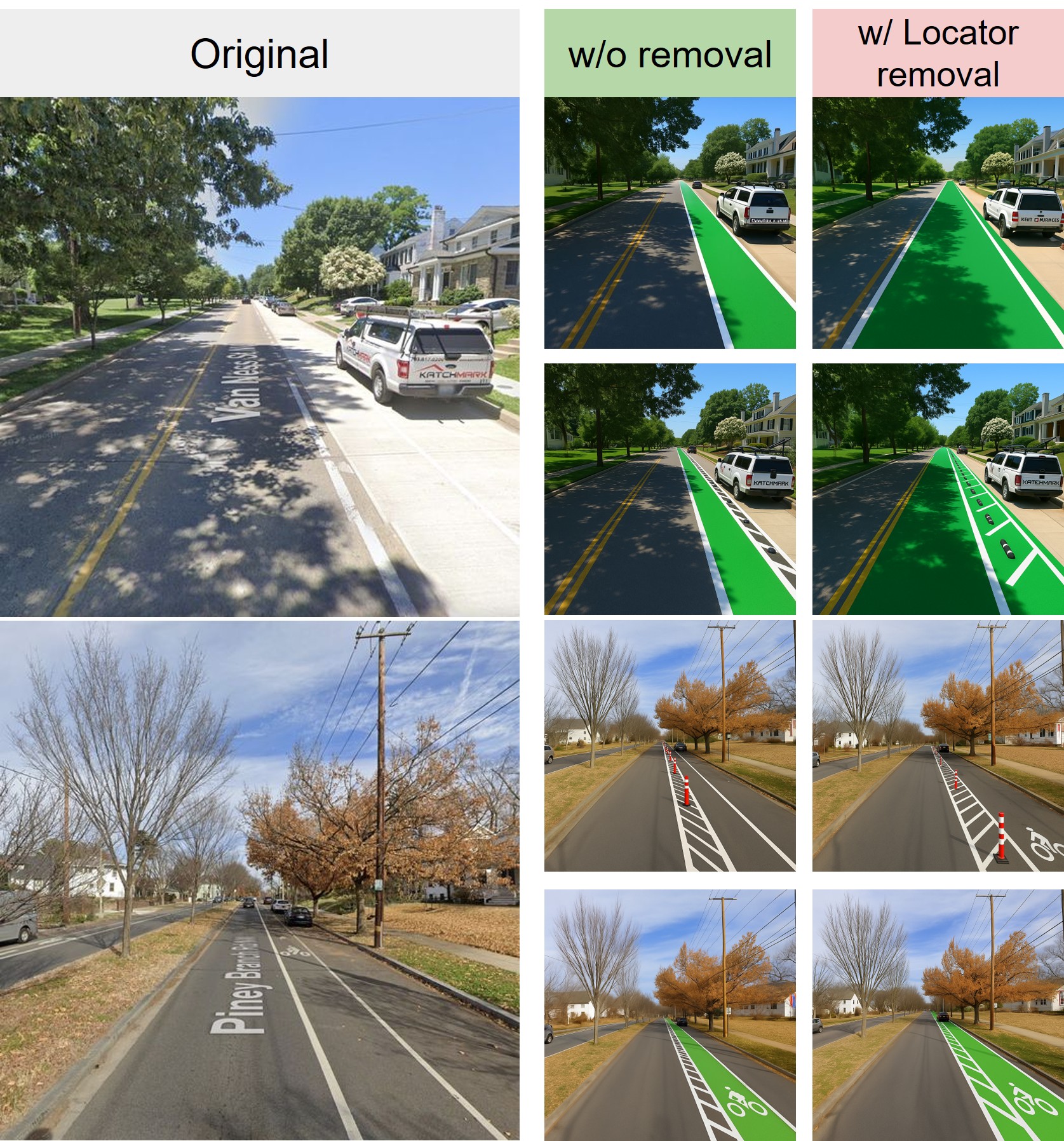}
    \caption{\textbf{Effect of removing the Locator Agent.} Each row shows the original street-view image (left), the output from the full pipeline (middle), and the output without the Locator Agent (right).}
    \label{fig:remove_locator}
\end{figure}

\paragraph{Removing the Locator Agent.}
Without the Locator Agent, the system loses reliable spatial grounding for where bicycle infrastructure should be edited or inserted. As shown in Fig.~\ref{fig:remove_locator}, the generator frequently misidentifies travel lanes as bike lanes and places surfacing or markings near the road center rather than adjacent to the curb. We also observe collateral edits to non-target road elements (e.g., partial rewriting of parking zones), indicating degraded background preservation. In contrast, the full pipeline produces curb-aligned lanes while preserving surrounding roadway features, highlighting the importance of explicit localization for geometry-correct and context-consistent design.

\begin{figure}
    \centering
    \includegraphics[width=0.75\linewidth]{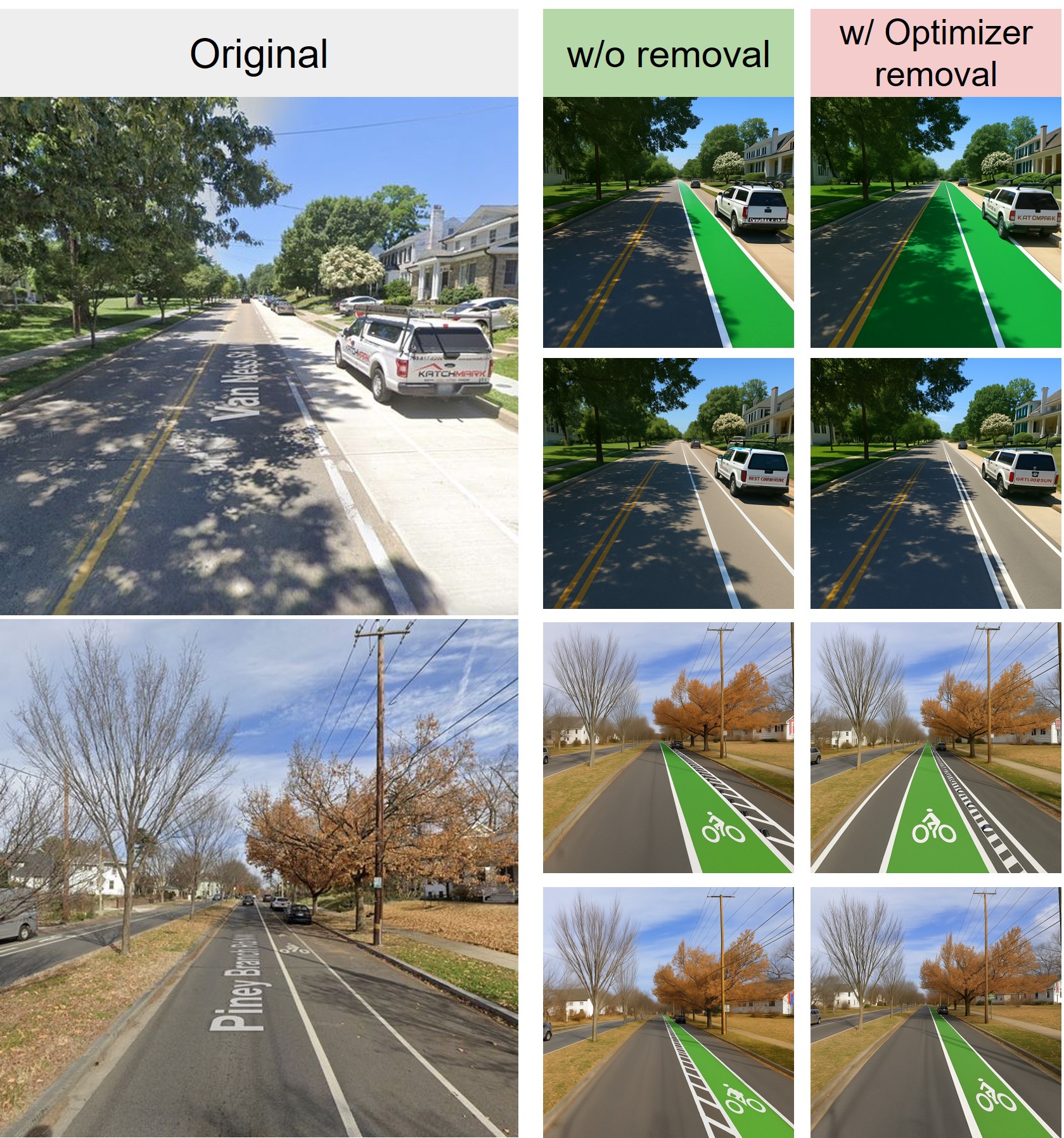}
    \caption{\textbf{Effect of removing the Prompt Optimization Agent.} Each row shows the original street-view image (left), the output from the full pipeline (middle), and the output without the Prompt Optimization Agent (right).}
    \label{fig:remove_optimizer}
\end{figure}

\paragraph{Removing the Prompt Optimization Agent.}
When the Prompt Optimization Agent is removed, the pipeline relies on raw user prompts, which are often under-specified for side, width, buffer type, symbol placement, and background-preservation constraints. Figure~\ref{fig:remove_optimizer} shows typical failures including mis-rendered boundaries (e.g., ``two parallel lines'' realized as unintended double boundaries), color spill beyond the intended lane region, omitted or flipped buffers/hatching, and misaligned symbols relative to roadway direction. These errors reflect the generator’s need to infer multiple latent specifications from ambiguous text. The Prompt Optimization Agent mitigates such failures by producing a structured, verifiable prompt grounded in the scene geometry and reinforced by in-context exemplars.

\begin{figure}
    \centering
    \includegraphics[width=0.75\linewidth]{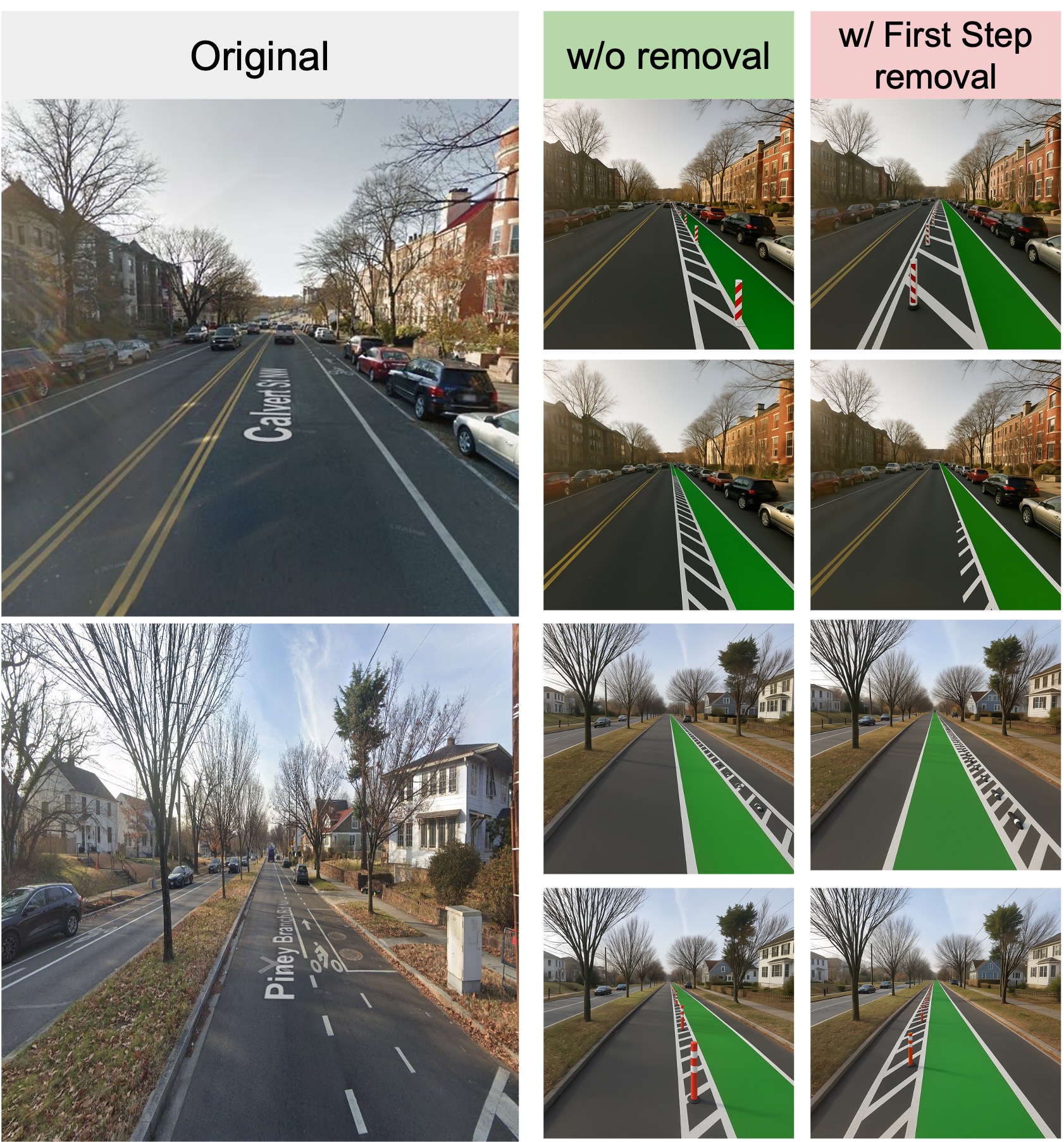}
    \caption{\textbf{Effect of removing the highlight-first step in the Design Generation Agent.} Each row shows the original street-view image (left), the output from the full pipeline (middle), and the output without the highlight-first step (right).}
    \label{fig:remove_step}
\end{figure}

\paragraph{Removing the highlight-first step.}
The two-step design generation strategy first creates a highlighted lane region and then renders the final infrastructure using that region as a spatial scaffold. When this highlight-first step is omitted (Fig.~\ref{fig:remove_step}), the generator must jointly solve localization, width control, and styling from text alone, which increases instability. We observe systematic width drift (overly wide/narrow or inconsistent tapering), spillover into parking/shoulder space, and perspective-inconsistent striping. Scenario-specific elements such as buffers and hatching are also more frequently omitted or misplaced. The highlight-first step provides an interpretable spatial prior that anchors the lane extent before styling is applied, thereby reducing hallucinations and improving compliance.

\begin{figure}
    \centering
    \includegraphics[width=0.95\linewidth]{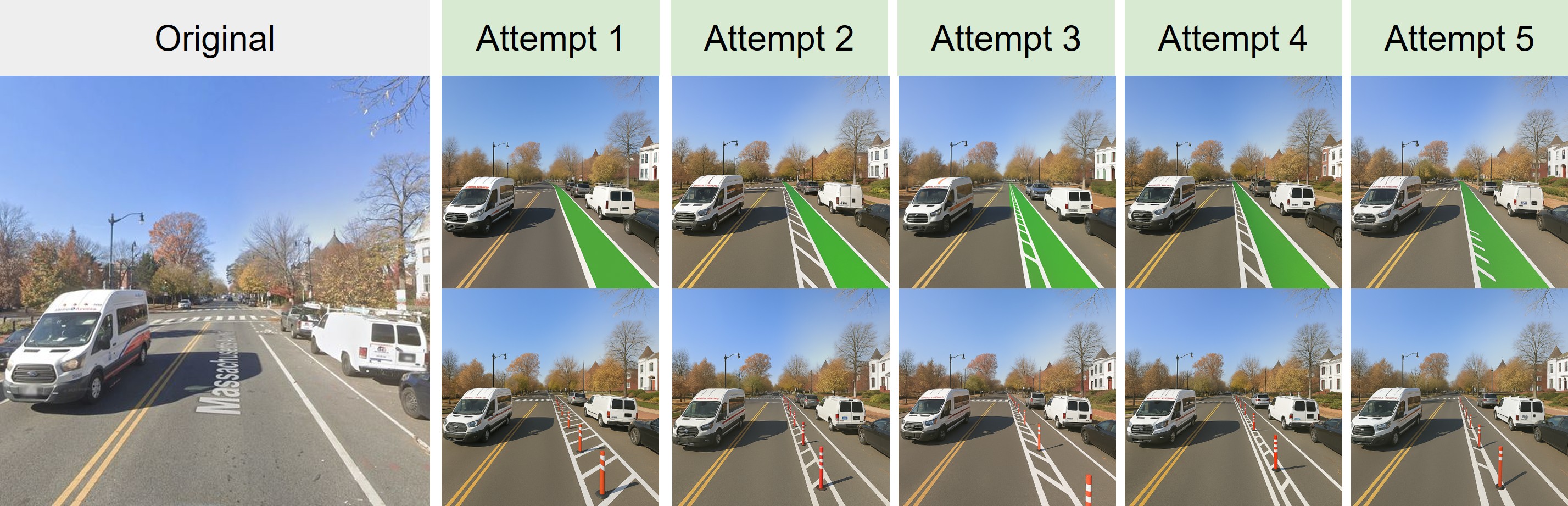}
    \caption{\textbf{Necessity of the Evaluator Agent.} Five independent generations from the same input image and prompt illustrate stochastic variation; the Evaluator Agent enables reliable selection of a compliant design from the candidate pool.}
    \label{fig:remove_evaluator}
\end{figure}

\paragraph{Necessity of the Evaluator Agent.}
Because current image generators are stochastic and do not enforce hard geometric constraints, repeated generations from the same input and prompt can vary substantially (Fig.~\ref{fig:remove_evaluator}). We observe variability in lane width, curb alignment, buffer/hatching presence, protective element spacing, and occasional color spill beyond the intended region; some samples violate the prescribed design scenario. To mitigate this sample-level variance, our pipeline generates a candidate pool and uses the Evaluator Agent to first re-rank candidates by lane-region similarity to a reference design and then verify prompt compliance via multimodal reasoning. Removing the Evaluator Agent would propagate inconsistent or noncompliant outputs, whereas its inclusion yields a stable selection of the most suitable candidate under noisy generation.

%% file: conclusion.tex
\section{Conclusion}

In this work, we present a multi-agent framework for bicycle infrastructure design that integrates advanced image generation models with reasoning MLLMs. The system decomposes the design process into several steps through specialized agents, enabling context-aware and spatially accurate modifications to street-view imagery. Both qualitative and quantitative evaluation results demonstrate that our approach can robustly produce accurate, contextually appropriate, and visually realistic bicycle infrastructure designs. By integrating advanced AI systems into the street design workflow, our method offers a promising tool to support bicycle infrastructure planning and facilitate design.

\section*{Limitations}
Despite the effectiveness of the proposed multi-agent framework, several limitations remain. First, the current system still cannot fully guarantee pixel-level accuracy in representing spatial relationships within the generated designs. While the generated bike lanes are generally aligned with the intended roadway regions, fine-grained positional accuracy is not always achieved, particularly in complex street layouts. Second, the correctness rate of a single generation pass remains relatively low, requiring multiple candidate generations before a satisfactory result is obtained. This increases computational cost and latency in the design workflow. Finally, the pipeline still involves a substantial degree of human intervention, especially manual image selection during data preparation. Reducing this reliance on human involvement is essential for improving automation and scalability in future work.

%% file: related_work.tex
\section{Related Work}
\vspace{-1em}
\subsection{AI-assisted Generative Design in Urban Planning}

Since its emergence, AI has advanced rapidly and found broad applications across diverse fields~\citep{bommasani2021opportunities, wang2024scalable, wang2023causality, wang2024near, li2024mosaic, li2025ruler, karpatne2025ai}, with recent efforts beginning to explore its potential in urban planning and design.
Wijnands et al.~\citep{wijnands2019streetscape} used unpaired GAN-based image translation on large Google Street View datasets stratified by population health, augmenting streetscapes to reveal actionable design cues—more green space, wider footpaths/sidewalks, fewer fences, and greater frontage compactness—informing healthier street and sidewalk design.
Ito et al. ~\citep{ito2024translating} quantifies the bias in car-mounted street-view imagery and introduces a GAN-based translation framework to convert road-center views into cyclist/sidewalk perspectives, aligning semantic indicators so bikeability/walkability assessments better inform bikeway and streetscape design.
Rajagopal et al.~\citep{rajagopal2023hybrid} similarly use CycleGAN to convert simple annotated road layouts into lifelike street-level images, enabling rapid prototyping of bike lane designs.
Calleo et al.~\citep{calleo2024exploring} employ a Real-Time Spatial Delphi method combined with GAN-based image synthesis to produce photorealistic street redesign visuals, significantly aiding stakeholder communication. 
Arrabi et al.~\citep{arrabi2025cross} propose a two-stage, geometry-preserving ground-to-aerial synthesis pipeline (BEV layout prediction from a street photo, then text-conditioned diffusion) to generate realistic overhead imagery that can aid transportation/streetscape planning. 
Zhang et al.~\citep{zhang2024urban} introduce a 3D diffusion approach conditioned on BEV layouts to generate large, unbounded urban scenes as semantic occupancy maps (and renderable images), enabling rapid what-if exploration of road-network design options. 
Hu et al.~\citep{hu2025ursimulator} develop a human-perception-guided prompt-tuning framework that locally edits street-view images with Stable Diffusion to simulate urban-renewal interventions and quantitatively boost perceived safety/beauty/liveliness—useful for previewing streetscape or bike-corridor upgrades.

\textbf{Collectively, these studies demonstrate the promise of generative AI  for urban and streetscape visualization, but they are ill-suited for bicycle-infrastructure design. }In particular, they still face notable limitations, such as insufficient spatial immersion and fine‐grained environmental cues to convey the on‐road cyclist experience, or high computational costs and labor‐intensive data preparation for generative models. In addition,\textbf{ the overall design workflow remains cumbersome, }requiring extensive manual adjustments and cross‐disciplinary coordination among human experts.

\subsection{Multi-agent system for Image Editing}

Recent developments in MLLMs and related multi-agent systems have driven substantial progress across a range of tasks~\citep{li2025caughtcheating,liang2025colorbench, chen2025musixqa,chen2025perceptions}, including image editing.
Gupta et al.~\citep{gupta2023visual} introduce VisProg, which demonstrated how an LLM-based single-agent planner could decompose intricate editing tasks, highlighting the benefits of agent-driven task segmentation but constrained by reliance on a single controller. 
Hang et al.~\citep{hang2025cca} proposed Collaborative Competitive Agents (CCA), employing two generators and a discriminator in an iterative feedback loop, where generators compete yet collaboratively improve via shared feedback, significantly enhancing robustness for complex multi-step edits.
Venkatesh et al. introduce CREA~\citep{venkatesh2025crea}, which employed distinct role-based agents (e.g., Creative Director, Art Critic) to iteratively refine images creatively, significantly outperforming single-prompt diffusion models by achieving greater output diversity and semantic alignment through a collaborative, human-like creative process.
Xie et al.~\citep{tianyidan2025anywhere} adopted a modular multi-agent approach for foreground-aware editing tasks, assigning dedicated agents for foreground semantics, object integrity, and background consistency. This system notably enhanced image quality and control compared to end-to-end models. EmoAgent~\citep{mao2025emoagent} tackled affective image manipulation by emulating cognitive painting workflows through planning, execution, and critique agents, significantly enhancing emotional expression and editing interpretability.
Marmot~\citep{sun2025marmot} employed specialized agents post-generation to correct object count, attributes, and spatial arrangement, substantially improving alignment with textual descriptions.  
Multi-Agent Collaboration-based Compositional Diffusion (MCCD)~\citep{li2025mccd} similarly used multimodal LLMs to parse prompts into object-specific agents, integrating regional outputs through hierarchical diffusion to achieve precise control and improved compositional consistency.  Role-specialized agents, LLM planners, and collaborative/competitive feedback reliably decompose complex edits, improve semantic alignment, and enhance robustness over single-prompt baselines. Building on this consensus, we tailor a coordinated agentic architecture to bicycle infrastructure scenario design, aligning agents with domain needs, so that modified scenario design to street-view imagery can be produced consistently.

%% file: supplement.tex
\section{Model Evaluation and Validation}

In this section, we justify our selection of the backbone model and assess the effectiveness of each component in the multi-agent generation pipeline through corresponding qualitative evaluations.

\subsection{Generation Backbone Comparison}
In this section, we compare our image generation backbone, GPT-image-1, with the state-of-the-art open-source image generation model, Stable Diffusion 3.5~\citep{stabilityai2025stable35}, demonstrating the rationale of backbone selection.
We focus on three design scenarios, Design Scenario 1, Design Scenario 6, and Design Scenario 7 (Please see Table~\ref{tab:boundaries_by_pattern}), because these designs are among the most frequently implemented in contemporary urban cycling infrastructure and are explicitly emphasized in widely used design guidelines such as the NACTO Urban Bikeway Design Guide~\citep{national2025urban}.  To ensure that observed performance differences are attributable solely to the generative models rather than content variability, we standardize the experimental conditions: for each design scenario, all models receive the exact same input street-view photograph and identical textual prompt. The only variable is the generative backbone, GPT-image-1 versus the three latest Stable Diffusion models, Stable Diffusion  3.5 variants (Stable Diffusion 3.5 Large, Stable Diffusion 3.5 Large Turbo, and Stable Diffusion 3.5 Medium), allowing us to directly attribute output differences to model-specific handling of spatial and semantic constraints.

\begin{figure}[h]
    \centering
    \includegraphics[width=1\linewidth]{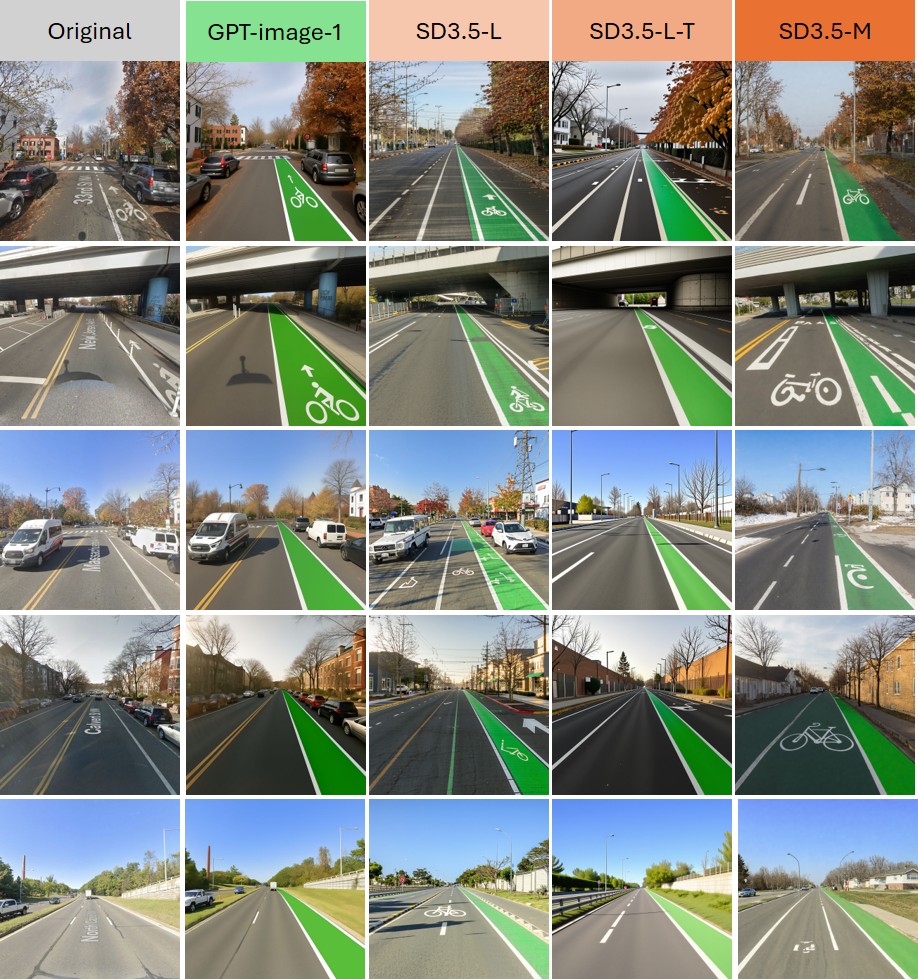}
    \caption{\textbf{Qualitative comparison of Desgin Scenario 1 outputs} generated by GPT-image-1 and three Stable Diffusion 3.5 variants (Stable Diffusion 3.5 Large (SD3.5-L), Stable Diffusion 3.5 Large Turbo(SD3.5-L-T), and Stable Diffusion 3.5 Medium(SD3.5-M).}
    \label{fig:comp_p1}
\end{figure}

\begin{figure}[h]
    \centering
    \includegraphics[width=1\linewidth]{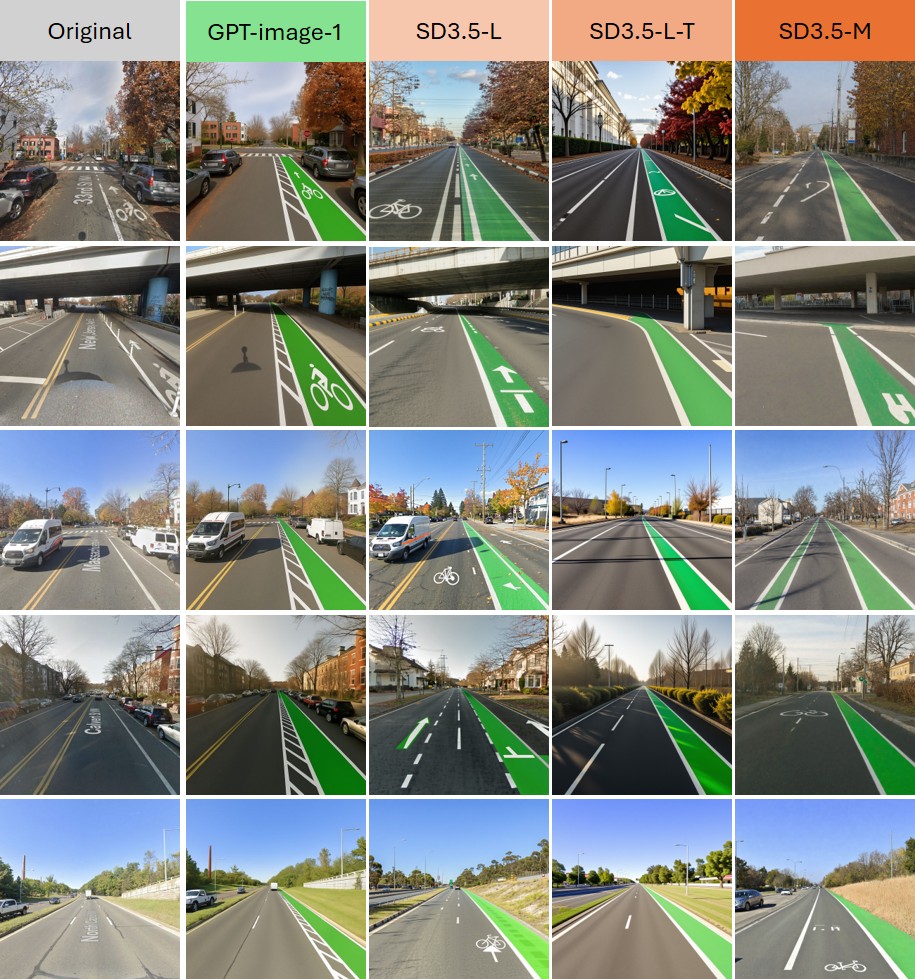}
    \caption{\textbf{Qualitative comparison of Desgin Scenario 6 outputs} generated by GPT-image-1 and three Stable Diffusion 3.5 variants (Stable Diffusion 3.5 Large (SD3.5-L), Stable Diffusion 3.5 Large Turbo(SD3.5-L-T), and Stable Diffusion 3.5 Medium(SD3.5-M).}
    \label{fig:comp_p6}
\end{figure}

\begin{figure}[h]
    \centering
    \includegraphics[width=1\linewidth]{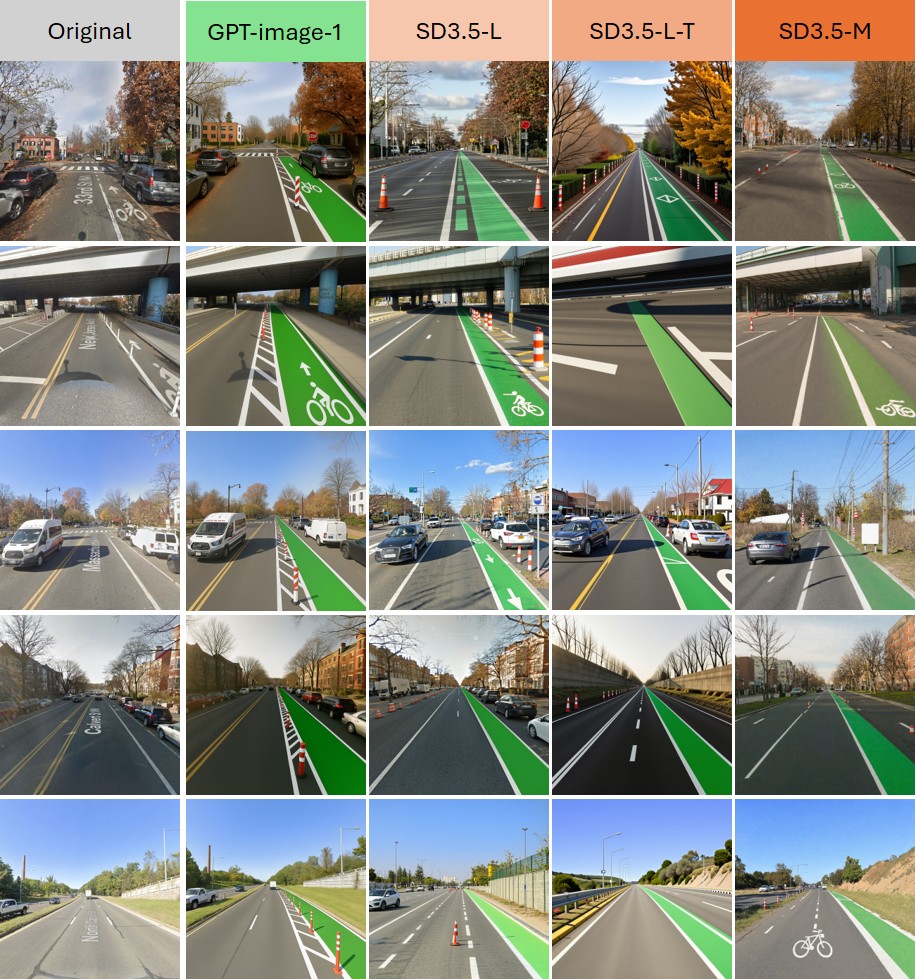}
    \caption{\textbf{Qualitative comparison of Desgin Scenario 7 outputs} generated by GPT-image-1 and three Stable Diffusion 3.5 variants (Stable Diffusion 3.5 Large (SD3.5-L), Stable Diffusion 3.5 Large Turbo(SD3.5-L-T), and Stable Diffusion 3.5 Medium(SD3.5-M).}
    \label{fig:comp_p7}
\end{figure}

Figures~\ref{fig:comp_p1},\ref{fig:comp_p6}, and~\ref{fig:comp_p7} present qualitative comparisons under identical input photographs and textual prompts, covering the three most commonly implemented bikeway typologies as emphasized in widely used design guidelines~\citep{national2025urban}.
Furthermore, to ensure that observed performance differences are attributable solely to the generative models rather than content variability, we standardize the experimental conditions: for each design scenario, all models receive the exact same input street-view photograph and identical textual prompt. From the qualitative comparison, we find that GPT-image-1 consistently adheres to the prompt with high fidelity, preserving all background elements while making precise, localized edits to the bicycle infrastructure. In Figure~\ref{fig:comp_p1}, the model modifies the current bike lane into a green-painted bike lane that follows road geometry and perspective, with bicycle glyphs correctly scaled, oriented, and positioned. In Figure~\ref{fig:comp_p6}, it generates a buffered lane of appropriate width and perspective without altering adjacent lane markings or roadside objects. In Figure~\ref{fig:comp_p7}, it accurately places a protected lane with bollards while maintaining occlusion relationships with visual elements. These results demonstrate strong semantic alignment between the textual description and visual output, with minor changes to unrelated scene content.

In contrast, the Stable Diffusion 3.5 variants (Large, Large Turbo, and Medium; SD3.5-L/SD3.5-L-T/SD3.5-M) exhibit recurrent failure modes despite visually realistic textures. The core limitation is twofold: these models neither reliably internalize where bicycle infrastructure must be placed relative to road elements, nor do they strictly follow our prompt to modify highly specialized, position-dependent details. Concretely, we observe (i) \textit{topological errors}: lanes crossing the centerline, appearing on the parking side, or breaking at junctions; (ii) \textit{metric/perspectival inconsistencies}: lanes rendered with an approximately constant image-space width rather than shrinking with depth, and buffers whose width varies along the road direction; (iii) \textit{symbolography mistakes}: bicycle glyphs and arrows with incorrect orientation, spacing, or lane offset; (iv)\textit{ occlusion/layering violations}: painted lanes rendered over vehicles or barriers instead of behind them; and (v)\textit{ unintended scene rewriting}: hallucinated buildings or altered road geometry, i.e., background replacement (cf. Figs.\ref{fig:comp_p1}, \ref{fig:comp_p6}, and\ref{fig:comp_p7}). These design scenarios suggest weak inductive bias for structured layout and scene topology: the models optimize global realism but lack mechanisms to enforce local, prompt-governed geometric constraints and strict background preservation demanded by infrastructure editing. Mitigation via Stable-diffusion-based fine-tuning is pragmatic but typically requires large, curated, domain-specific datasets and task-specialized conditioning signals, which are resource-intensive and, for this narrowly scoped editing task, have conflict over our research motivation~\citep{zhang2023adding}. Accordingly, we center our evaluation on GPT-image-1 and do not extend to additional Stable-diffusion-based variants.

%% file: prompt.tex
\section{Prompts}
\label{sec:prompt}

\begin{figure*}[h]
  \centering
  \parbox{1\textwidth}{
        \rule{1\textwidth}{1.5pt} 
        Prompt of Locator Agent\\
        \rule{1\textwidth}{0.8pt} 

        \textbf{System Prompt} \\
        You are a helpful vision assistant to identify the bike lane from the image and describe its location accurately.\\[4pt]

        \textbf{User Prompt} \\
        Your task is to describe precisely the physical location and boundaries of the primary bike lane shown on the right side of the roadway in the provided image in sentences. Typically, bike lanes are defined clearly by white lines on both sides; however, boundary variations exist, and it is possible that one side of the bike lane may instead be marked by a buffer zone (e.g., a painted area with diagonal stripes), a curb, sidewalk edge, or physical separators like bollards or raised barriers. If either the left or right boundary is a buffer zone or another physical separator, treat that separator as part of the bike lane in your description. In this task, you should specifically detail that, in most cases, the primary bike lane is bordered by two parallel white lines: one white line forming the left boundary separating it from motor-vehicle lanes, and another white line forming the right boundary separating it from parking cars, sidewalks, or curbs. Clearly note the exact placement relative to these adjacent roadway features.\\

        \rule{1\textwidth}{0.8pt} 
  }
\caption{
Prompt used in Locator Agent to request \textbf{GPT-o3} to describe the physical location and boundaries of the bicycle infrastructure from an image.
}
\label{prompt:bike_lane_location}
\end{figure*}

\begin{figure*}[h]
  \centering
  \parbox{1\textwidth}{
        \rule{1\textwidth}{1.5pt} 
        Prompt of Prompt Optimization Agent\\
        \rule{1\textwidth}{0.8pt} 

        \textbf{System Prompt} \\
        You are an expert prompt optimizer. Your job is to transform a draft prompt about depicting or updating bicycle infrastructure in roadway images into a precise, unambiguous, self-contained instruction for an image-generation or editing model. Preserve the user's intent exactly; remove ambiguity; standardize terminology; and keep constraints measurable and consistent. Explicitly specify: (a) lane position on the right-hand side of the road; (b) lane width with units; (c) surface color policy (green vs. standard road surface); (d) left and right boundaries. Define the left boundary as the continuous solid white line separating the bike lane from motor-vehicle lanes; define the right boundary as either a continuous solid white line or a clearly marked buffer zone (diagonal white stripes) and, if present, any physical separators (e.g., bollards) should be described as part of the boundary, not within the lane. If the user forbids green paint, clearly say “No green paint.” If the user requires a fully green lane, require that the green area is strictly contained between two continuous solid white lines. Do not invent parameters not present in the user input; when information is missing, keep statements general while matching the style of the examples. Write in clear imperative voice. Output only the optimized prompt—no commentary, headings, or quotes.\\[6pt]

        \textbf{User Prompt} \\
        \#\#Example 1\#\# \\
        The area currently painted green with two white boundary lines represents the existing bike lane. Your task is to clearly depict an updated bike lane that is approximately 6 feet wide, fully painted green, strictly contained between two prominent, continuous, solid white boundary lines. Ensure these white boundary lines clearly define the left and right edges of the green bike lane area, positioned along the right-hand side of the road. Do not allow any green paint to extend beyond the white boundary lines.\\[6pt]

        \#\#Example 2\#\# \\
        The area currently painted green with two white boundary lines represents the existing bike lane. Clearly depict an updated bike lane approximately 6 feet wide, located along the right-hand side of the road. Do not paint the updated bike lane green; use the standard road surface color only. Clearly mark both boundaries of the bike lane: 1) Left boundary: a prominent, continuous solid white line. 2) Right boundary: a clearly marked narrow buffer zone adjacent to the bike lane, filled with prominent diagonal white stripes, and bounded on both sides by solid white lines. Ensure the updated bike lane is clearly defined by the solid white lines on both sides, distinctly separate from the striped buffer zone on its right side. No green paint should be applied.\\[6pt]

        \#\#Example 3\#\# \\
        The area currently painted green with two white boundary lines represents the existing bike lane. Clearly depict an updated bike lane approximately 4 feet wide, located along the right-hand side of the road. Do not paint the updated bike lane green; use the standard road surface color only. Clearly mark both boundaries of the updated bike lane: 1) Left boundary: a prominent, continuous solid white line. 2) Right boundary: a clearly marked narrow buffer zone adjacent to the bike lane, filled with prominent diagonal white stripes, bounded on both sides by solid white lines, and distinctly featuring vertical red-and-white striped bollards placed at regular intervals. Ensure the updated bike lane is clearly defined by the solid white lines on both sides, distinctly separate from the striped buffer zone and bollards on its right side. No green paint should be applied.\\[10pt]

        Rewrite the \textit{\{USER\_PROMPT\}} into a single optimized prompt that follows the System Prompt and mirrors the style of the In-Context Examples. Preserve all stated constraints; make boundary definitions explicit (left vs. right); avoid contradictions; and keep the output concise (preferably $\leq$ 130 words). Output only the optimized prompt.\\

        \rule{1\textwidth}{0.8pt} 
  }
\caption{
In-context learning prompt of Prompt Optimization Agent.
}
\label{prompt:icl_bike_prompt_optimization}
\end{figure*}

\begin{figure*}[h]
  \centering
  \parbox{1\textwidth}{
        \rule{1\textwidth}{1.5pt} 
        Prompt of auxiliary step in Design Generation Agent\\
        \rule{1\textwidth}{0.8pt} 

        \textbf{System Prompt} \\
        You are a helpful assistant for precise roadway image editing. Follow instructions exactly, maintain visual realism (perspective, lighting, shadows), and avoid adding elements not requested. Preserve the legibility of traffic-control devices and do not alter objects unless explicitly instructed.\\[6pt]

        \textbf{User Prompt} \\
        Edit the entire existing bike-lane corridor on the right side of the road into a \{COLOR\}-painted lane. Treat the corridor as the current bike-lane surface plus any immediately adjacent buffer zones or physical separators that belong to the lane configuration. Keep the lane strictly contained between two continuous solid white boundary lines (left and right). Apply the following: \\[4pt]
        \quad-- Boundaries: ensure both left and right boundaries are prominent, continuous solid white lines that follow roadway curvature. Do not let any \{COLOR\} paint cross, touch, or bleed over these lines.\\
        \quad-- Buffer zones: if a narrow striped buffer exists adjacent to the lane, replace its interior stripes with a uniform \{COLOR\} infill, but retain the solid white lines that bound the buffer as the lane’s outer edges.\\
        \quad-- Physical separators: keep bollards, armadillos, curbs, or raised barriers intact and unpainted; they should remain visually above the \{COLOR\} base and aligned along the boundary. Do not recolor or remove them.\\
        \quad-- Exclusions: exclude painted street names, arrows, lane labels, and crosswalk markings from recoloring. Do not extend the \{COLOR\} paint into motor-vehicle lanes, parking spaces, sidewalks, or curbs.\\
        \quad-- Consistency: preserve road texture and lighting, respect occlusions (vehicles, pedestrians), and keep the lane’s original footprint (do not widen or narrow).\\
        \quad-- Output: deliver a clean, continuous \{COLOR\}-painted lane on the right side of the road, strictly bounded by solid white lines, with all exclusions observed.\\

        \rule{1\textwidth}{0.8pt} 
  }
\caption{Prompt of auxiliary step in Design Generation Agent}
\label{prompt:highlight_area_paint}
\end{figure*}

\begin{figure*}[h]
  \centering
  \parbox{1\textwidth}{
        \rule{1\textwidth}{1.5pt} 
        Prompt of Evaluator Agent\\
        \rule{1\textwidth}{0.8pt} 

        \textbf{System Prompt} \\
        You are a strict binary evaluator for roadway images. Examine the candidate image (and the provided reference image, if any) and decide whether it shows a bike lane located on the right side of the road that satisfies \emph{all} features listed in the User Prompt. Minor visual variations (e.g., perspective, lighting, small width deviations) are acceptable only if each required feature is clearly present and recognizable. If any required feature is missing, ambiguous, occluded, or contradicted, respond \texttt{no}.\\[4pt]
        Output format: a single lowercase word, exactly \texttt{yes} or \texttt{no}. Do not add punctuation, spaces, explanations, or any other text.\\[6pt]
        
        \textbf{User Prompt} \\
        Answer ONLY \texttt{yes} or \texttt{no}:\\[4pt]
        Does the image show a bike lane on the right side of the road with the following key features? Minor variations are allowed, but all features should be clearly recognizable: \\[4pt]
        \textbf{1. Left Boundary:}\\
        \quad-- Narrow buffer zone adjacent to the bike lane.\\
        \quad-- Buffer zone bounded by solid white lines on both sides.\\
        \quad-- Prominent diagonal white stripes filling the buffer zone.\\
        \quad-- Rounded, semi-flexible rubber lane dividers (“armadillos”) placed centrally and evenly spaced within the buffer zone. Dividers should be dome-shaped, black with white reflective stripes.\\[4pt]
        \textbf{2. Right Boundary:}\\
        \quad-- Prominent continuous solid white line marking the right-hand edge of the bike lane.\\[6pt]
        The image should closely match the reference image provided, clearly depicting both boundary conditions. Answer \texttt{no} if these conditions are not sufficiently met.\\

        \rule{1\textwidth}{0.8pt} 
  }
\caption{An example of a prompt for evaluation used in Evaluator Agent.}
\label{example:evaluator_bikelane_binary}
\end{figure*}